\documentclass[conference]{IEEEtran}
\IEEEoverridecommandlockouts
\usepackage{cite}
\usepackage{comment}
\usepackage{amsmath,amsfonts}
\usepackage{array}
\usepackage[caption=false,font=normalsize,labelfont=sf,textfont=sf]{subfig}
\usepackage{textcomp}
\usepackage{stfloats}
\usepackage{url}
\usepackage{verbatim}
\usepackage{graphicx}
\usepackage{array}
\usepackage{caption} 
\usepackage{xcolor}
\usepackage{multirow}
\usepackage[ruled, lined, linesnumbered, longend]{algorithm2e}
\usepackage{soul} 

\def\BibTeX{{\rm B\kern-.05em{\sc i\kern-.025em b}\kern-.08em
    T\kern-.1667em\lower.7ex\hbox{E}\kern-.125emX}}
    
\SetKwInput{KwOutput}{Output}   
\SetKwInput{KwInput}{Input}

\newcommand{\fixme}[1]{{\textcolor{blue}{\em\bf{[FIXME: #1]}}}}

\begin{document}



\title{Reinforcement Learning-based Knowledge Graph Reasoning for Explainable Fact-checking}

\author{
\IEEEauthorblockN{Gustav Nikopensius}
\IEEEauthorblockA{\textit{Institute of Computer Science} \\
University of Tartu, Estonia\\
gustav.nikopensius@ut.ee}
\and

\IEEEauthorblockN{Mohit Mayank}
\IEEEauthorblockA{\textit{Outplay, India} \\
mohit.m@outplayhq.com}
\and

\IEEEauthorblockN{Orchid Chetia Phukan}
\IEEEauthorblockA{\textit{Department of CSE} \\
IIIT-Delhi, India\\
orchidp@iiitd.ac.in}

\and
\IEEEauthorblockN{Rajesh Sharma}
\IEEEauthorblockA{\textit{Institute of Computer Science} \\
University of Tartu, Estonia\\
rajesh.sharma@ut.ee}
}

\maketitle

\begin{abstract}
Fact-checking is a crucial task as it ensures the prevention of misinformation. However, manual fact-checking cannot keep up with the rate at which false information is generated and disseminated online. Automated fact-checking by machines is significantly quicker than by humans. But for better trust and transparency of these automated systems, explainability in the fact-checking process is necessary. Fact-checking often entails contrasting a factual assertion with a body of knowledge for such explanations. An effective way of representing knowledge is the Knowledge Graph (KG). There have been sufficient works proposed related to fact-checking with the usage of KG but not much focus is given to the application of reinforcement learning (RL) in such cases. To mitigate this gap, we propose an RL-based KG reasoning approach for explainable fact-checking. Extensive experiments on FB15K-277 and NELL-995 datasets reveal that reasoning over a KG is an effective way of producing human-readable explanations in the form of paths and classifications for fact claims. The RL reasoning agent computes a path that either proves or disproves a factual claim, but does not provide a verdict itself. A verdict is reached by a voting mechanism that utilizes paths produced by the agent. These paths can be presented to human readers so that they themselves can decide 
whether or not the provided evidence is convincing or not. This work will encourage works in this direction for incorporating RL for explainable fact-checking as it increases trustworthiness 
by providing a human-in-the-loop approach. 

\end{abstract}



\section{Introduction}

Social media platforms are not only used by billions for communication but has become an alternative platform (compared to TV, print media) to receive news \cite{Boczkowski2017IncidentalNH}. These platforms have also being misused for spreading misinformation \cite{Flintham2018,jagtap2021misinformation} and its various forms such as fake news \cite{mayank2022deap,dhawan2022game}, rumors \cite{butt2022goes,sharma2021identifying}, misleading news \cite{sharma2022mis}, etc. 
With the sheer pace of misinformation generated manual fact-checking can not keep up. So, automated fact-checking 
must be employed to tackle the menace of misinformation. One of the most prominent ways is through the usage of machine learning (ML) techniques \cite{islam2020deep,shakshi2023mis}. 
The majority of such techniques handle the context of the misinformation for its detection. For example, investigating what kind of users propagate the misinformation \cite{ghenai2018fake} or processing the content with natural language processing (NLP) techniques to derive clues from the way the misinformation is written, etc. \cite{su2020motivations}. 

It is important to mention that the rise in misinformation has bring forth the focus on fact-checking as it helps in controlling the spread of misinformation \cite{martin2022facter,tymoshenko2021strong,sharma2022facov}. 
 Providing explanations for fact-checking tools leads to trust among the people and offer better usability. Examples of the works related to explanations include formulating the task of explanation generation as an extractive-abstractive summarization task using transformer models \cite{Atanasova2020, Kotonya2020}. These particular works deal with specific domains such as political news \cite{Atanasova2020} and public health \cite{Kotonya2020}. In addition, to use these approaches, the datasets used in these studies have been annotated with gold-standard justifications by journalists. This approach is not scalable as annotation is a costly task. Alternatively, Knowledge graphs (KGs) have also been employed to provide a mechanism in which reliable third-party sources such as scientific articles and Wikipedia are used for fact-checking \cite{Ahmadi2019, GadElrab2019, Schatz2021}.  However, the use of reinforcement learning (RL) in such cases hasn't garnered much attention.

In this work, we propose an RL-based KG reasoning approach for explainable fact-checking that employs an RL agent to produce explanations in the form of relevant knowledge from the KG in the form of a path. The paths are arrived at by a technique from the field of KG reasoning known as multi-hop reasoning where an RL agent hops over facts in an attempt to connect two nodes 
that correspond to real-world entities. The path describes the relation between those two entities 
and this path is used to classify a statement as true or false and also provides a human-readable explanation of the classification result. 

To summarize, the main highlights of this work are as follows:
\begin{enumerate}
    \item To the best of our knowledge, this is the first work that has used a multi-hop reasoning agent for explainable fact-checking. We use two KGs (FB15K-277 and NELL-995) that are commonly used in multi-hop reasoning tasks to evaluate the performance of the proposed model. The KGs represent real-world relationships, sets of true facts are gathered from these KGs, and corresponding subsets of false facts are generated to simulate false facts.
    \item The results show that a multi-hop reasoning model can successfully check facts and provide highly human-readable explanations for the classifications. The agent is also able to provide explanations and a basis for classification which is not directly tied to any specific fact-checking task.
\end{enumerate}

The rest of the paper is organized as follows. Section \ref{related} discusses the related works. Next,  in Section \ref{methodology} we describe the methodology of the proposed approach, followed by the experiments and its results in Section \ref{experiments}. We conclude in Section \ref{conclusion} with some future directions. 
\section{Related Work}\label{related}
This section describes previous works related to multi-hop reasoning (Section \ref{sec:mh}) and fact-checking with KG (Section \ref{sec:fc}). 

\subsection{Multi-hop reasoning}\label{sec:mh}
KG reasoning infers new knowledge from existing knowledge in the KG, and detects errors in the KG. Large scale KGs are inherently incomplete. The main effort in KG reasoning research is to ``fill in'' the missing knowledge. This is done by either KG traversal known as multi-hop reasoning \cite{Das2017,  Lv2020, Wan2021} 
or by mining logic rules based on KG semantics and using those rules for reasoning \cite{Zhang2019, Qu2021, Minervini2021}.  
There are two main directions in multi-hop reasoning: entity prediction and relation prediction. The former looks for a tail entity for a head entity and relation pair
and the latter looks for a missing relation between two entities.
The task of looking for a missing tail entity is more similar to our work of finding a path that leads from a given head entity to the desired (true) tail entity. The major difference is that the proposed approach receives the claimed tail entity as an input as well.

Reasoning in the continuous vector space of KG embeddings was first proposed by Xiong et al. \cite{Xiong2017}. Compared to Das et al. \cite{Das2017a} 
which rely on Path Ranking Algorithm (PRA) to produce the initial paths, Xiong et al. \cite{Xiong2017} utilizes a policy-based RL algorithm REINFORCE to learn paths in the KG. Das et al. \cite{Das2017} builds on Xiong et al. \cite{Xiong2017} by directing the task from relation prediction to link prediction, a more complicated task, and also building a model that can handle multiple reasoning entities. 
We use both the proposed target entity and relation as input, thus differing from both Das et al. 
\cite{Das2017} as well as Xiong et al. \cite{Xiong2017}.

\subsection{Fact-checking with Knowledge Graph}\label{sec:fc}
Ciampaglia et al. \cite{Ciampaglia2015} framed the problem of fact-checking as finding a connection between the subject and object of a fact statement: assertion is true if there is an edge or short path between the entities in a KG, and false otherwise. Their proposed method makes use of a truth valuation function that accounts for the length of the path between the two entities as well as the generality of intermittent entities, where generality means \textit{how many statements it participates in.}
Shiralkar et al. \cite{Shiralkar2017} builds upon the notion of generality of intermittent entities but considers multiple paths between the subject and object entities as a knowledge stream.
These approaches rely on pre-extracted subgraphs. Inversely, the proposed approach saves resources by integrating relevant subgraph extraction as a part of the explainable classification process.

Previous research is also involved with mining the KG for patterns to be used in classifying a fact statement.
Shi et al. \cite{Shi2016} mined the KG for patterns associated with specific relations. 
Lin et al. \cite{Lin2019} used a supervised pattern discovery algorithm to find patterns relevant to training data. Lin et al. \cite{Lin2018a} builds upon this work by also taking into account the ontological closeness of entities. 
 Unlike these works, this work wholly operates in the continuous vector space and is not formally concerned with patterns, although patterns do arise naturally from the RL aspect.
\par

\noindent \textbf{Explainable fact-checking with Knowledge Graphs: }Ahmadi et al. \cite{Ahmadi2019} used KGs to assess claims and accompany these with human-readable explanations. This is achieved through a combination of logical rule discovery and probabilistic answer set programming (ASP). The KG is mined for rules in the form of triples associated with specific relations and these relations' negations. For fact assessment the heads of these rules are replaced by claim values and the KG is scavenged for corresponding triples. Explanations and an assessment are derived from these triples through ASP. Gad-Elrab et al. \cite{GadElrab2019} used rules to extract explanations from KGs and text resources. Schatz et al. \cite{Schatz2021} mined for patterns for fact-discovery in the biomedical domain to produce human-readable explanations to professionals for drug or treatment discovery. The main difference between these previous approaches and the proposed  approach is that it does not rely on mined patterns but a learned policy and KG embeddings.

\section{Methodology}\label{methodology}
\begin{figure}
    \centering
    \resizebox{0.5\textwidth}{!}{\includegraphics{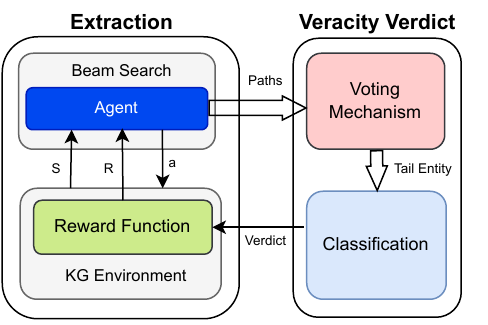}}
    \caption{Proposed Framework; ``S'' stands for State which contains the input claim and the path that is built starting from the head entity, ``R'' stands for the Reward which is used to update the agent's policy function at the end of the episode, ``a'' stands for the action which the Agent sends to the environment} 
    \label{fig:framework}
\vspace{-5pt}
\end{figure}

This section provides an overview of the proposed framework for explainable fact-checking (Figure \ref{fig:framework}). 
We first explain the path extraction process (Section \ref{evi})  succeeded by the veracity verdict (Section \ref{ver}), and the environment in Section \ref{env}.

\subsection{Evidential path extraction}
\label{evi}
The path extraction process follows multi-hop reasoning practices in which a policy-based agent sequentially extends an inference path. In contrast to traditional multi-hop reasoning tasks where one part of the claim is hidden from the agent, the proposed framework presents the agent with the entire claim and, in addition, the inference path, as it is being formed, as observations. We boost the quantity and versatility of extracted evidential paths by combining the policy-based agent with the beam search algorithm. 
 
\noindent \textbf{Policy-based agent: } The policy-based agent receives a claim as input and, in return, extracts a path from the KG. The path extraction process follows the pattern of a Markov decision process (MDP). The agent considers an observation, comprising the input claim and current path, to decide upon an action with which to extend the current path, after which the path is updated, and the agent receives an updated observation. The process is repeated until the expiration of a predetermined number of time steps.
The process results in a path that begins with the claim's head entity and bears, 
after sufficient training, a semantic similarity to the claim. And thus, the resulting path can be used to judge upon the veracity of the input claim. As semantically similar, the tail entities of the claim and the evidential paths should match for a true claim and differ for a false claim.
The policy function maps the observation to a probability distribution over the action space which comprises all of the unique relations within the KG. The distribution is then filtered to only include relations that connect the current tail entity of the path. This is done to only include relations that can be used to extend the path. A \textit{self-loop} relation is included within the action space in case the agent decides to remain at the current tail entity. 
The policy function is parametrized by a neural network to maximize the expected reward for the path-based classification of the claim.

\noindent\textbf{Beam search: }
The beam search algorithm allows us to increase the number of paths which are extracted during an episode. 
It is customary to guide the relation choice in KG reasoning tasks, but to leave the entity choice to random chance. In this work, 
we propose a heuristic function that takes into account the fitness of the entities which the relations lead to by leveraging the KG scoring function \cite{trouillon2016complex}. 
By doing this, the search algorithm takes into account how fitting the resulting entity is as a correct tail entity for the evidential path by scoring it as the tail entity in the input triple. As the result of the scoring function is less important in the early steps of path extension, this function is the base of an exponential function, where the step number is the exponent. This function is added to the probability calculated for a given action by the policy function and the sum is used to compare viable path extension options to each other in the beam search process. The final equation stands (Equation \ref{eq1}): 

\begin{equation}
\label{eq1}
\pi(a|s, \theta) + Re(< r, e_{s}, \bar{e}_c >)^{step}
\end{equation}
where, $\pi(a|s, \theta)$ is the output probability of the policy function for the action a, given state s and policy function parameters $\theta$, and $Re(< r, e_{s}, \bar{e}_c >)$ 
is the scoring function and $step$ is the step of the MDP episode. Here, $r$, $e_{s}$, and $\bar{e}_c$ are vector representations of relation, head entity, and tail entity. The tuple signifies the dot product of the three vectors. $Re$ means the real part of the dot product.

\subsection{Veracity verdict}\label{ver}
The veracity verdict is arrived upon by grouping the extracted paths and counting all unique path tail entities. The straightforward approach would be to choose the most popular tail entity as the choice tail entity to be compared to the claim tail, but since the result can be a tie, a heuristic weighting is added to the votes. The beam search heuristic for the final step of each path is used to weight the vote of each entity. This practically excludes the possibility of a tie between the contending tail entities and adds leverage to the entities which the agent is more confident in and which received a higher score from the KG triple scoring function which is used to evaluate whether a given triple belongs to the KG or not and is used to fit vector embeddings to entities and relations.
The verdict is then evaluated against the ground truth and the result is used to update the agent which extracted the paths.

\subsection{Environment}\label{env}

The environment is modeled as a Markov decision process (MDP) which controls the interaction between the RL agent (the Agent) and the knowledge graph. A 4-tuple $\mathcal{<S, A, T, R>}$ represents the MDP. Each element is described below.

\noindent \textbf{State ($\mathcal{S}$):} The state consists of entities and relations. Two representations of the state are operated with in parallel: 
1) a human-readable collection of real-world elements: entities and relations, and 
2) a corresponding collection of embeddings - vector representations that capture the semantic information of the state elements. The state contains the fact claim, traversed path, and current location of the agent which is itself an entity. The initial state consists of an input triple which is the fact claim. 
Thus the initial state can be thought of constituting as, 
$<$ \textit{claimed Target}, \textit{claimed Relation}, \textit{beginningOfPath}, ...$>$

The embedded state is padded with zero vectors since the state is used as input to the policy network which requires a fixed-size input. These zero vectors are incrementally replaced by relation-entity embedding pairs during the agent's path traversal process. The final entity in the set of these pairs is of particular importance: this is the entity that is used to give a veracity verdict for the fact claim. So in essence, once the path traversal is completed then the veracity verdict stands as: $$claimedTarget == finalEntity$$

\noindent\textbf{Action ($\mathcal{A}$): }
Not only is the input size of the policy network fixed but so is the output size - it is always the same size as the action space. The policy network takes the state as input and computes probabilities for each action in the action space. This can be thought of as the agent making an estimate: which actions are most likely to take me from the current entity to the desired final entity?
During training time, one of the top three most probable actions are chosen for each transition. The choice is random with a distribution that is calculated from the output of the policy network to fit the new sample size. 
Random choice is introduced in order to reduce the likelihood of the agent getting stuck looking for sub-optimal paths brought forth by the local minima problem. 

\noindent\textbf{Transition ($\mathcal{T}$): }
Choosing a new action entails a transition of state. The transition itself is a function $T:$ $S X A$ $->$ $S$, which means that performing an action at a given state produces a new state. This new state is the input triple, a traversed path plus the choice action and a new current entity. The new current entity is a node in the KG that is connected to the previous current entity by the choice action. It is not uncommon for an entity to have multiple neighbors that are all connected to it via the same relation. 
Most previous literature concerning multi-hop reasoning picks a neighboring entity at random \cite{Xiong2017, Das2017}, the focus being on the performance of the RL agent, unsuited for the task of choosing a candidate from a variable size set of entities.

\noindent\textbf{Reward ($\mathcal{R}$): }
The agent is given a reward once the path traversal is complete. The reward is 1 if the final entity in the found path is equal to the true target.
Otherwise, the reward is 0.
\[
    reward = 
\begin{cases}
    1,& \text{if final entity = true target}\\
    0,              & \text{otherwise}
\end{cases}
\]

\section{Experiments}\label{experiments}

Here, we first describe the datasets (Section \ref{subsec:data}) followed by details regarding implementation and hyperparameters (Section \ref{implement}), results of our experimentations in Section \ref{sec:resl}, explainability of the proposed approach (Section \ref{explain}), and lastly a discussion of the results of our experiments (Section \ref{discuss}).

\subsection{Dataset}\label{subsec:data}

The experiments are conducted on two widely used benchmark datasets: NELL-995 \cite{Xiong2017} and FB15K-237 \cite{Toutanova2015}. Table \ref{table:KGstats} describes the specifics pretraining to the two datasets. NELL-995 is generated from the 995th iteration of the NELL system \cite{Carlson2010} by selecting the triples with the top 200 most frequent relations. FB15K-237's facts are created from FB15K \cite{Bordes2013} with redundant relations removed. Both datasets contain inverse triples to aid with path finding, allowing the reasoning agent to step backward in the graph. 
We chose three reasoning tasks per dataset: \textit{origin}, \textit{tvLanguage}, \textit{nationality} for FB15K-237, and \textit{athletePlaysInLeague}, \textit{athletePlaysForTeam}, \textit{worksFor} for NELL-995.
To evaluate the agent's performance, each reasoning task contains approximately ten negative examples for each positive sample, and the negative samples were generated by replacing a true claim's tail with a faked entity.

\begin{table}[h!]
\centering
\caption{Statistics of the datasets under consideration; \#Ent, \#Rel, and \#Fact denotes the number of unique entities, relations, and triples respectively}\label{Tbl:datatable}
\begin{tabular}{|l|c|c|c|}
\hline 
Dataset   & \#Ent & \#Rel & \#Fact  \\ \hline
FB15k-237 & 14,505                        & 237   & 272,115 \\ \hline
NELL-995  & 75,492                        & 200   & 154,213 \\ \hline
\end{tabular}
\label{table:KGstats}
\end{table}

\subsection{Implementation Details and Hyperparameters} \label{implement}

\noindent \textbf{Knowledge Graph Embeddings: }We use \textit{ComplEx} \cite{Trouillon2016} embedding model from \textit{Ampligraph} \cite{ampligraph} library for KG embeddings. The model's hyperparameters are chosen to balance model complexity and training time. We set the batch size as 50 and train the model for 1000 epochs. The embedding dimension is set to 20. We use Adam optimizer with a learning rate of 1e-4 and negative log-likelihood loss function. 
To prevent overfitting, we use an L3 regularization term with a regularization strength of 1e-5. We also set the random seed to 0 to ensure the reproducibility of the results. 

\noindent \textbf{Policy Network: }We use REINFORCE algorithm to train the policy network and \textit{PyTorch} library for the implementation. It comprises of ReLU as an activation function in the intermediate layers and Softmax in the output layer to generate a probability distribution over the actions given the current state. The network has hidden state size of 128 neurons and was initialized using the Xavier initialization method. 
We train the network for 100,000 episodes 
for each reasoning task, each consisting of 3 steps. We use  \textit{Adam} 
optimizer with a learning rate of 0.001. To encourage exploration, we randomly chose one of the top three most likely actions as the action to take in each state during training. \par

During the evaluation, we use beam search to obtain multiple likely actions according to the policy network's probability distribution for each state. This allows us to explore multiple possible paths through the action space and obtain a more robust estimate of the policy's performance. Three beam numbers, 3, 5, and 10 
are considered. We assess the performance of our model using an 80:20 train-test split. 
We use hits@k metric, which is commonly used in knowledge graph reasoning tasks. 
 The hits@k metric measures the proportion of test samples where the correct answer is among the top k-ranked answers returned by the model. In addition, we use accuracy as a metric for the voting mechanism. This was calculated as the proportion of test samples where the weighted majority vote was the correct answer. We acknowledge that the varying degrees of information in the knowledge graphs are dependent on alternative evidential paths. However, setting thresholds to consider the number of alternative paths for each reasoning task is complex and requires further research. As such, we did not take this into account in our evaluation.

\subsection{Experimental Results}\label{sec:resl}

\noindent \textbf{Basis for interpreting the results: } The two KGs used in experimentation are different, particularly in terms of their density features, with FB15K-237 containing 18.8 triples per entity and NELL-995 containing two triples per entity. This difference may affect the model's accuracy. 
The subsets for both KGs consist of 3 single relation tasks, with each KG also having a combined subset containing all three relations. This diversity in task assignment may complicate the agent's task, but it seems to actually assist the agent by giving meaning to the input triple's relation. In single relation tasks, the input triple's relation never changes and therefore does not inform the agent's actions. \par


\noindent \textbf{Results: } 
Tables \ref{Tbl:searchfb} and \ref{Tbl:votingfb} show the results of our proposed approach on Hits@k and Voting Accuracy for FB15K-237. Similarly, Tables \ref{Tbl:searchnl} and \ref{Tbl:votingnl} show the results of our proposed approach on Hits@k and Voting Accuracy for NELL-995.
The results show that the method is able to achieve high accuracy in finding a path to the true target entity with a high enough beam number, as indicated by the results of the hits@k metric. A higher beam number means more unique paths are considered, increasing the probability of reaching the target entity. Table II shows that considering more paths increases the proportion of input triples for which at least one path reached the target entity. But that's not the case always, considering more paths can even decrease accuracy because it requires the agent to discard more paths as is the case for \textit{origin} in Table \ref{Tbl:votingfb}, and \textit{athleteplaysinleague} and \textit{worksfor} in Table \ref{Tbl:votingnl}. 
However, the accuracy drops slightly after the voting process, as the correct path must be selected from the paths found. In particular, the "origin" dataset in FB15K-237 (Table \ref{Tbl:votingfb})
 has shown to be anomalous, achieving lower accuracy than expected despite being a larger dataset. The voting task has proven to positively impact accuracy, especially for the combined dataset in FB15K-237, indicating the model's scalability. Nonetheless, the results demonstrate that voting on an answer reduces accuracy, as the correct path and target entity may be correctly identified but still be undermined by a higher majority of an entity in the vicinity of the source entity. Both FB15K-237 and NELL-995 follow similar trends in terms of results regarding beam numbers and an accuracy trade-off in voting accuracy.
\textc

\begin{table}[h!]
\centering
\caption{Hits@k for FB15K-237; acc stands for accuracy} 
\label{Tbl:searchfb}
\begin{tabular}{|c|c|c|c|c|}
\hline
            & \multicolumn{1}{l|}{dataset size} & \multicolumn{1}{l|}{acc., b=3} & \multicolumn{1}{l|}{acc., b=5} & \multicolumn{1}{l|}{acc., b=10} \\ \hline
origin      & 852                                                             & 0.464                                                      & 0.53                                                       & 0.661                                                       \\ \hline
tvLanguage  & 614                                                             & 0.52                                                       & 0.659                                                      & 0.894                                                       \\ \hline
nationality & 7134                                                            & 0.866                                                      & 0.907                                                      & 0.921                                                       \\ \hline
combined    & 8600                                                            & 0.807                                                      & 0.886                              & 0.943                               \\ \hline
\end{tabular}
\label{tabel:fb15kresults}
\end{table}

\begin{table}[h!]
\centering
\caption{Voting accuracy for FB15K-237; acc stands for accuracy} 

\label{Tbl:votingfb}
\begin{tabular}
{|c|c|c|c|c|}

\hline FB15K-237   & dataset size & acc., b=3 & acc., b=5 & acc., b=10 \\ \hline
origin      & 852                & 0.357         & 0.369                                 & 0.292                                  \\ \hline
tvLanguage  & 614                & 0.463         & 0.472                                 & 0.642                                  \\ \hline
nationality & 7134               & 0.727         & 0.75                                  & 0.765                                  \\ \hline
combined    & 8600               & 0.252         & 0.602         & 0.727          \\ \hline
\end{tabular}
\end{table}

\begin{table}[h!]
\centering
\caption{Hits@k for NELL-995; acc stands for accuracy}
\label{Tbl:searchnl}
\resizebox{\columnwidth}{!}{\begin{tabular}
{|c|c|c|c|c|}
\hline
NELL-995     & dataset size & acc, b=3 & acc, b=5 & acc, b=10 \\ \hline
athleteplaysinleague & 12660              & 0.919         & 0.971                                 & 0.994                                  \\ \hline
athleteplayssport    & 6781               & 0.903         & 0.853                                 & 0.868                                  \\ \hline
worksfor             & 9669               & 0.748         & 0.78                                  & 0.825                                  \\ \hline
combined             & 29110              & 0.485         & 0.575         & 0.611          \\ \hline
\end{tabular}}
\end{table}

\begin{table}[h!]
\centering
\caption{Voting accuracy for NELL-995; acc stands for accuracy}
\label{Tbl:votingnl}
\resizebox{\columnwidth}{!}
{\begin{tabular}{|c|c|c|c|c|}
\hline
NELL-995     & dataset size & acc, b=3 & acc, b=5 & acc, b=10 \\ \hline
athleteplaysinleague & 12660              & 0.858         & 0.938                                 & 0.875                                  \\ \hline
athleteplayssport    & 6781               & 0.871         & 0.828                                 & 0.83                                   \\ \hline
worksfor             & 9669               & 0.604         & 0.59                                  & 0.509                                  \\ \hline
combined             & 29110              & 0.45          & 0.467         & 0.464          \\ \hline
\end{tabular}}
\end{table}

\subsection{Explainability}

\label{explain}

For the purpose of providing explanations, we present examples of single-relation tasks. 
 Each example is evaluated based on the proposed approach's performance in arriving at the correct conclusion, the informativeness of the path, and the ground truth interpretation. We provide six examples in total.


\noindent\textbf{Example 1: KG contains outdated information}
\\Claim: Dick Cheney works for retailstore Halliburton
\\Path: Dick Cheney – leads organization → retailstore Halliburton

\noindent The agent has found a correlation between a person leading an organization and working for the organization. This is an interesting successful case, since Dick Cheney stopped leading Halliburton the day George W. Bush took him on as vice president candidate for the 2000 US presidential elections. This means the presented ground truth is actually false due to outdated information in the KG. 
\newline
\noindent\textbf{Example 2: Inconclusive path}
\\Claim:	Coach John Gruden works for sportsteam Buccaneers.
\\Path:	Coach John Gruden $\leftarrow$organization hired person– sportsteam Buccaneers

\noindent The agent attempts to find a solution by considering the relation “organization hired person”.
In fact Tampa Bay Buccaneers had hired John Gruden in 2002 and Gruden worked for the team until 2008.
Thus, the claim that John Gruden works for the Buccaneers was true from 2002-2008, but is not true anymore, meaning  the agent's reasoning is good, but the conclusion arrived at is false, indicating the need for further investigation.
The agent has learned a correlation between the relations “worksfor” and “organizationhiredperson inv” meaning if an organization has hired a person, then the person works for that organization.
\newline
\textbf{Example 3: Informative path with complex reasoning}
\\Claim:	Coach Brendan Shanahan plays Hockey.
\\Path:	Brendan Shanahan –plays for team→ the Devils –team plays sport→ Hockey

\noindent This is a good example of the agent finding a path that does not just contain a (near) synonym for the claimed relation (like \textit{works for} and \textit{leads organization}), but finding a more complicated path.
The agent finds that Brendan Shanahan plays for the Devils hockey team, and then finds out that the Devils play hockey.
\newline
\noindent\textbf{Example 4: Agent stays on the fence}
\\Claim:	Journalist Amir Taheri works for Los Angeles County.
\\Path:	Amir Taheri → Amir Taheri 
\newline
The Agent is unable to move beyond the source entity, resulting in an inconclusive path. This may be possible because of three reasons:
\begin{enumerate}
    \item The length of the path is fixed to three relations, no more, no less.
    \item The agent must have the possibility to decide upon a target entity and “stick with it”
    \item The agent can decide to stay put at an entity for all three transitions.
\end{enumerate}
Since most resultative paths only take one or two hops from entity to entity, the action of staying put is very common. Thus it’s always one of the top actions to choose for any state. With beam search also in place, it is very common that one path will result in the agent staying put for the entirety of the path.
\newline
\textbf{Example 5: Two true paths with only one interpreted as correct}
\\Claim: Ben Folds originates from Nashville.
\\Path 1: Ben Folds --place of birth$\rightarrow$Winston-Salem
\\Path 2: Ben Fold --location$\rightarrow$ Nashville

These paths state that Ben Folds was born in Winston-Salem and that he is located in Nashville. For this specific case, the ground truth is that Ben Folds originates from Winston-Salem, so the first path leads to the true target and the second path reinforces the "false claim".
The majority of explainable paths for the "origin" subset consist of one of two (near) synonymous relations: "place of birth" and "location".
It is also common for one of the path relations to lead to a correct target entity and the other to lead to an incorrect one, the meaning of "origin" staying quite ambiguous. The subset is concerned with musical artists, so sometimes "origin" will mean the artists place of birth, other times it will mean the city in which they became established as a musician which will often coincide with the "location" relation.
\newline
\textbf{Example 6: Two true explainable paths with only one being considered as correct}
\\Claim: Kylie Minogue originates from Australia.
\\Path 1: Kylie Minogue –place of birth→ Melbourne
\\Path 2: Kylie Minogue –place of birth→ Melbourne ←contains– Australia

\noindent The claim that Kylie Minogue originates from Australia is true in real life, but in regard to the ground truth, it is considered false.
Kylie Minogue originates from Australia which is true, but the ground truth is that Kylie Minogue originates from Melbourne, which is a city in Australia. Due to the setup of the veracity verdict mechanism, a path that concludes that Kylie Minogue originates from Australia is be considered false even though the path contains Melbourne and is very much explainable.

\subsection{Discussion}
\label{discuss}
The results of this study demonstrate that selecting an entity to be evaluated for veracity has a trade-off with accuracy. While the search algorithm may have identified the correct entity, it may not have been selected by the voting mechanism. This trade-off is not linear, and tasks with lower accuracy in identifying a correct path are more severely impacted by the results of voting. Analysis of common patterns in the paths generated by the agent highlighted several issues that should be considered in future research on this problem.

Firstly, it is crucial to use an up-to-date KG to ensure the accuracy of fact classifications. Outdated information in a KG can produce false results. Secondly, the verdict mechanism would benefit from identifying and incorporating inconclusive paths. The agent's motivation to move away from the source entity should also be increased. Additionally, paths that do not end with the target entity but support the ground truth should also be utilized to arrive at a correct conclusion. Lastly, the classification process could be improved by learning to deal with ambiguous relationships such as ``origin''. These findings suggest several avenues for future research to improve the accuracy and efficiency of veracity verdicts.
\section{Conclusion}\label{conclusion}
This work proposes a combination of reinforcement learning and KG to provide explainable fact classifications. The problem was formatted as a multi-hop reasoning problem enhanced by a beam search-based voting mechanism, which uses the paths produced by a graph hopping reinforcement learning agent to vote on the veracity of a claimed fact statement.
The results show that this sort of approach can be used for fact classification with varying accuracy dependent on the nature of the input claims, and it produces highly human-readable explanations for the classifications.
Future directions will make use of more massive KGs and extract facts-to-check from fake news datasets to provide a highly usable framework for real-world misinformation detection tasks.

\section{Acknowledgment}
This work has received funding from the EU H2020 program under the SoBigData++ project (grant agreement No. 871042), by the CHIST-ERA grant No. CHIST-ERA-19-XAI-010,  (ETAg grant No. SLTAT21096), and partially funded by HAMISON project. 

\bibliographystyle{IEEEtran}
\bibliography{main}

\end{document}